\def\year{2022}\relax
\documentclass[letterpaper]{article} 
\usepackage{aaai22}  
\usepackage{times}  
\usepackage{helvet}  
\usepackage{courier}  
\usepackage[hyphens]{url}  
\usepackage{graphicx} 
\usepackage{natbib}  
\usepackage{caption} 
\DeclareCaptionStyle{ruled}{labelfont=normalfont,labelsep=colon,strut=off} 
\usepackage{algorithm}
\usepackage{amsmath}
\usepackage{amsfonts}
\usepackage{subcaption}
\usepackage{bbm}
\usepackage{arydshln}
\usepackage{wasysym}
\usepackage[algo2e,ruled,vlined,linesnumbered]{algorithm2e}

\usepackage{booktabs} 
\usepackage{subcaption}
\usepackage{bbm}
\usepackage{arydshln}
\usepackage{wasysym}
\usepackage{xcolor}

\usepackage{newfloat}
\usepackage{listings}
\floatstyle{ruled}
\newfloat{listing}{tb}{lst}{}
\floatname{listing}{Listing}

\title{Information-Theoretic Bias Reduction via Causal View of Spurious Correlation}
\author{
Seonguk Seo$^{1}$ \quad Joon-Young Lee$^{2}$  \quad Bohyung Han$^1$ 
}
\affiliations{
 $^1$ECE \& ASRI \& GSAI, Seoul National University 
\\ $^2$Adobe Research
\\  {\tt\small \{seonguk, bhhan\}@snu.ac.kr \quad jolee@adobe.com}
}

\newcommand{\etal}{\textit{et al}.}
\newcommand{\ie}{\textit{i}.\textit{e}.}
\newcommand{\eg}{\textit{e}.\textit{g}.}

\begin{document}

\maketitle


\begin{abstract} 
We propose an information-theoretic bias measurement technique through a causal interpretation of spurious correlation, which is effective to identify the feature-level algorithmic bias by taking advantage of conditional mutual information.
Although several bias measurement methods have been proposed and widely investigated to achieve algorithmic fairness in various tasks such as face recognition, their accuracy- or logit-based metrics are susceptible to leading to trivial prediction score adjustment rather than fundamental bias reduction.
Hence, we design a novel debiasing framework against the algorithmic bias, which incorporates a bias regularization loss derived by the proposed information-theoretic bias measurement approach.
In addition, we present a simple yet effective unsupervised debiasing technique based on stochastic label noise, which does not require the explicit supervision of bias information.
The proposed bias measurement and debiasing approaches are validated in diverse realistic scenarios through extensive experiments on multiple standard benchmarks.

\end{abstract} 

\section{Introduction}

Various recognition algorithms based on deep neural networks have achieved remarkable performance improvement by learning useful patterns from a large number of training examples, and have started to be deployed in many real-world applications.
However, the objective of the optimization problem is mainly concerned about the final accuracy, which makes trained models vulnerable to unexpected decision rules affected by spurious correlation.
Although such decision rules work well on most of training examples, they often lead to poor worst-case generalization performance and make learned models unfair to the examples in under-represented groups.
For instance, a few existing works~\cite{kim2016data, buolamwini2018gender} have discovered the performance gap across demographic subgroups in real-world face recognition tasks; the accuracy of a model on \emph{darker-skinned women} is often much lower than that on \emph{lighter-skinned men}.
This tendency becomes a major weakness in achieving algorithmic fairness and generalizing on unseen test environments with domain or distribution shifts.

Mitigating spurious correlation has recently emerged as an important issue in learning debiased model~\cite{hardt2016equality, zhao2017men, zhang2018mitigating, li2019repair, wang2019iccv, gong2020jointly, sagawa2020investigation, GroupDRO}.
Given the target and bias variables, they typically suppress spurious correlation by making the algorithm independent of the bias variables while maintaining the predictions to the target variables.
To this end, various measurements about algorithmic fairness or bias have been introduced, \eg, demographic parity~\cite{calders2009building}, equality of odds~\cite{hardt2016equality}, and group-fairness accuracy~\cite{GroupDRO, zhang2020adaptive}.
Although they have been widely used to maintain fairness, their accuracy- or logit-based bias measurement schemes may lead to inaccurate algorithmic bias measurement.
For example, we observe that fine-tuning the last linear classification layer of a model is sufficient to achieve high unbiased accuracy and worst-group accuracy, even without updating the biased representations.

To better quantify algorithmic bias, we first formulate the spurious correlation from a causal point of view.
Following the strict definition of fairness, we derive the condition of independence between a bias variable and a predicted target variable, which is evaluated by their mutual information.
However, in our causal view, it gives a biased estimation of spurious correlation.
To handle the issue, we propose a new bias measurement technique based on \textit{conditional} mutual information using feature representations, which is claimed to quantify the algorithmic bias more accurately while maintaining the original objective, fairness.

Based on the new bias measurement, we propose a debiasing framework to mitigate the algorithmic bias by augmenting a loss term for bias regularization, which is derived by the conditional mutual information.
We also introduce a simple yet effective unsupervised debiasing technique by exploiting stochastic label noise, which also prevents a model from capturing spurious correlation even without the prior knowledge of bias information.
Our experiments verify that both approaches are helpful for alleviating the algorithmic bias while preserving model accuracy.

The main contributions of our work are summarized as follows.
 \begin{itemize} 
	\item[$\bullet$] 
	 We propose an information-theoretic measurement technique for the amount of algorithmic bias via conditional mutual information on learned feature representations, which is derived from the causal view of spurious correlation.

	 \item[$\bullet$] Based on the new bias measurement approach, we propose two novel debiasing frameworks; one employs a information-theoretic bias regularization loss and the other is with stochastic label noise even without the supervision of bias information.
	 
	 \item[$\bullet$] We evaluate our bias measurement scheme and debiasing techniques in various realistic scenarios and achieve promising results on multiple standard benchmarks.  
\end{itemize}


\section{Setup and Preliminaries}
\label{sec:setup} 
Let $(X, Y, Z) \in \mathbb{R}^m \times \mathbb{R} \times \mathbb{R}$ be a triplet representing a joint distribution over the space $(\mathcal   {X}, \mathcal{Y}, \mathcal{Z})$, where $X$ is an input variable, $Y$ is a target variable, and $Z$ is a bias variable.
We denote a learned model by a function $c \in \mathcal{C}$ mapping $\mathcal{X}$ to $\mathcal{Y}$.
The classification function $c: X \rightarrow \hat{Y}$ consists of the feature extractor $g: X \rightarrow F$ and the linear classifier $h: F \rightarrow \hat{Y}$, \ie, $c(\cdot) = h(g(\cdot))$, where $F \in \mathbb{R}^d$ is a feature representation and $\hat{Y} \in \mathcal{Y}$ is the predicted target variable.
Following the conventional notations in probability theory, we will use the comma ($,$) to denote the joint distribution and the semicolon ($;$) to separate the input arguments of mutual information.
For example, $I(X;Y,Z)$ indicates the mutual information between $X$ and the joint distribution of $Y$ and $Z$.

\subsection{Group-Fairness Accuracy}
\label{sec:fairness_acc}
In classification tasks, unbiased accuracy, worst-group accuracy, or accuracy disparity are employed to address non-uniform accuracy and evaluate algorithm fairness and robustness in the presence of dataset bias~\cite{GroupDRO, zhang2020adaptive, wilds2021, zhang2020towards}.
Formally, if $Y_i \in \{1, ..., A\}$ and $Z_i \in \{1, ..., B\}$ denote the target and bias values of $X_i$, respectively, then the unbiased accuracy is given by
\begin{align}
\frac{1}{AB} \sum_{a, b} \frac{\sum_{i}{\mathbbm{1}(c(X_i)= Y_i = a, Z_i=b)}}{\sum_{i}{\mathbbm{1}(Y_i=a, Z_i=b)}}, 
\label{eq:unbiased_acc}
\end{align}
which indicates the average accuracy over all groups defined by a pair of target and bias values.
Other metrics, the worst-group accuracy and the accuracy disparity, are obtained by the worst accuracy of all groups and the discrepancy between the best and the worst group, respectively.

However, the unbiased accuracy may not be a good metric for evaluating the amount of bias in data by itself.
Given a baseline model converged sufficiently with \textit{hair color} classification on the CelebA dataset, we fix its feature extractor $f(\cdot)$ and fine-tune its classification layer~$h(\cdot)$ with a simple resampling technique to reduce class imbalance problem.
We select \textit{gender} as a bias variable for evaluation.
Although the features in this model remain unchanged even after the fine-tuning phase, it attains a significant performance gain, 8\% points in the unbiased accuracy and 19\% points in the worst-group accuracy.
This result implies that the accuracy- or logit-based bias measurement methods may not be able to capture the innate bias residing in the features, and that high fairness scores can be achieved by adjusting prediction scores in linear classification layers.

\subsection{Estimating Mutual Information}
To identify the feature-level bias, we first introduce the mutual information, which measures the co-dependence between two variables.
For random variables $X$ and $Y$ over the space $\mathcal{X} \times \mathcal{Y}$, the mutual information is defined as
\begin{align}
    I(X; Y) = H(X) - H(X|Y) = H(Y) - H(Y|X),
\end{align}
where $H(\cdot)$ denotes the Shannon entropy.
The mutual information is also defined by the Kullback-Leibler (KL) divergence between the joint distribution of two random variables and the products of their marginal distributions:
\begin{align}
\label{eq:mi_kl}
    I(X; Y) &= D_{KL}(P_{(X,Y)}||P_X \otimes P_Y),
\end{align}
which implies that the larger the divergence between the joint and the product of the marginals is, the stronger the dependence between $X$ and $Y$ is.

However, its exact computation is tractable only for discrete variables or continuous variables under some constraints.
Recently, Belghazi~\etal~\cite{belghazi2018mutual} propose a neural estimator of mutual information (MINE) between continuous high-dimensional random variables, by offering a lower bound of the mutual information based on the Donsker-Varadhan representation\footnote{The Donsker-Varadhan representation provides the KL-divergence as a supremum over all functions $T$: $D_\text{KL}(P\parallel Q) = \sup_{T:\Omega\rightarrow \mathbb{R}} \mathbb{E}_P[T]  - \log(\mathbb{E}_Q[\exp(T)])$. }~\cite{donsker1975asymptotic},
\begin{align}
    I(X; Y) \geq I_\phi(X; Y) &= \sup_{\phi\in \Phi} \mathbb{E}_{P_{(X,Y)}}[f_\phi(x, y)]   \\ &- \log (\mathbb{E}_{P_{X}\otimes{P_Y}}[\exp(f_\phi(x, y))]), \nonumber
    \label{eq:mine}
\end{align}
where $f_\phi(\cdot, \cdot)$ is a statistical neural network parameterized by $\phi \in \Phi$.
The expectations $\mathbb{E}_{P_{(X,Y)}}$ and $\mathbb{E}_{P_{X}\otimes{P_Y}}$ are approximated using empirical sampling from the joint distributions and the products of the marginal distributions, respectively.
By maximizing the right-hand side of \eqref{eq:mine} with respect to $\phi$, we can obtain a tighter lower bound, which leads to a more accurate estimation of the mutual information.


\section{Cobias: Bias Measurement with Conditional Mutual Information}
\label{sec:biasmeasure} 

\begin{figure*}[t]
    \centering
    \begin{subfigure}[b]{0.47\linewidth}    
        \centering
        \includegraphics[height=4.cm]{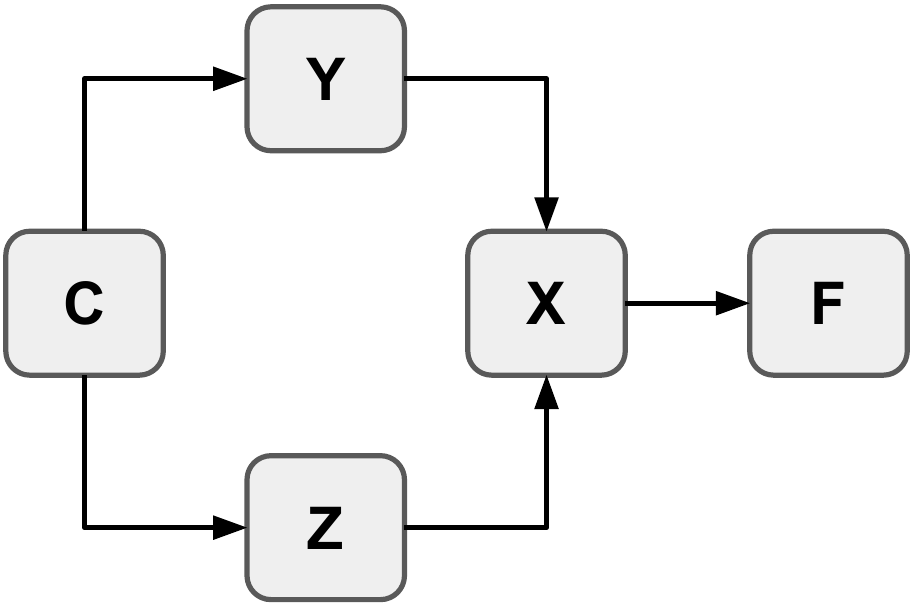}
        \caption{The diagram of structural causal model. We aim to measure the co-dependence between $Z$ and $F$ via $Z\rightarrow X\rightarrow F$, but it is tricky because there exists a backdoor path $Z\leftarrow C\rightarrow Y\rightarrow X\rightarrow F$.}
        \label{fig:diagramA}
    \end{subfigure}
    \hspace{0.4cm}
    \begin{subfigure}[b]{0.47\linewidth}      
        \centering
        \includegraphics[height=4.2cm]{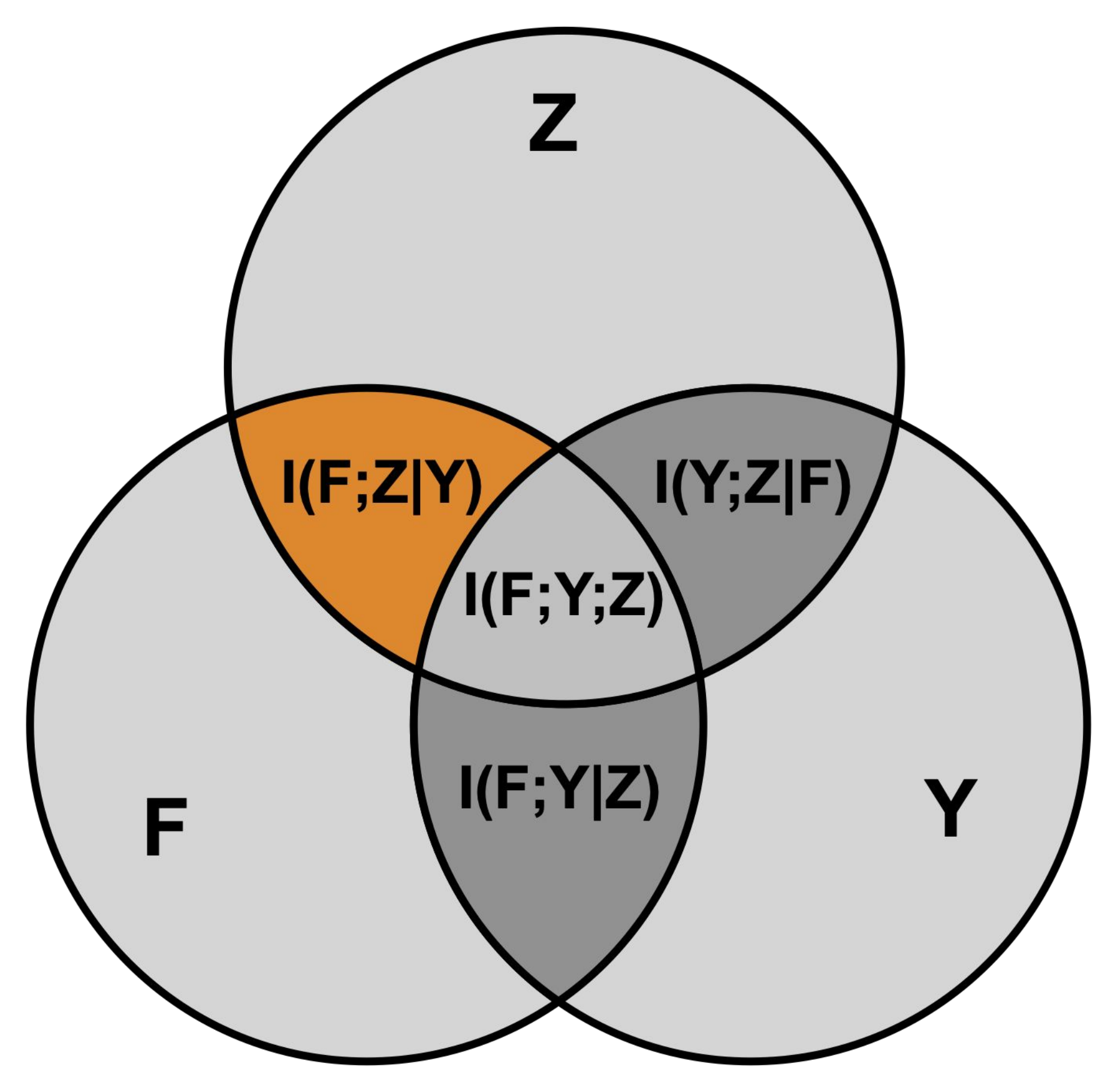}
        \caption{Venn diagram of information-theoretic measures for $F, Y,$ and $Z$.
        The region in orange denotes the mutual information between $F$ and $Z$, excluding the amount of information that explains $Y$.
        }
        \label{fig:diagramB}
    \end{subfigure}
    \caption{The causal and information-theoretic diagrams of our problem setting. }
    \label{fig:diagram}
\end{figure*}

We interpret the spurious correlation in a causal view, and present the bias measurement technique based on the conditional independence derived from the causal view.

\subsection{Causal Model of Spurious Correlation}
\label{subsec:causal_model}
We formulate the causal relationship among input image $X$, target label $Y$, bias variable $Z$, context prior $C$, and feature representation $F$ using a structural causal model~\cite{spirtes2000causation}.
Figure~\ref{fig:diagramA} illustrates the relationship, where the direct link $A \rightarrow B$ indicates that $A$ is the cause of $B$.
Note that we adopt the feature vector $F$, instead of logits or predicted target values as discussed before.

From the viewpoint of information leakage~\cite{dwork2012fairness}, it is possible to quantify the algorithmic bias by measuring the co-dependence between feature and bias variables via mutual information, \eg,  $I(Z; F)$.
However, as depicted in Figure~\ref{fig:diagramA}, the path between $Z$ and $F$ contains not only the direct path $Z\rightarrow X\rightarrow F$, but also a backdoor path $Z\leftarrow C\rightarrow Y\rightarrow X\rightarrow F$.
Because the model is trained to maximize the mutual information between $Y$ and $F$ via $Y\rightarrow X\rightarrow F$, the backdoor path is constructed naturally if there exists the correlation between $Y$ and $Z$.
To focus on the direct path, we should block the backdoor path, which is easily done by conditioning on $Y$.
Since the path $C\rightarrow Y\rightarrow X$ is a serial connection, conditioning on $Y$ blocks the backdoor path and we can now measure the mutual information between $Z$ and $F$ via the direct path only.

\subsection{Cobias: Conditional Mutual Information as Bias Measurement}
\label{sec:cobias}
Generalizing the mutual information to multivariate cases, we derive the conditional mutual information with three variables $X$, $Y$, and $Z$, which is given by
\begin{align}
    I(X; Y|Z) &= \mathbb{E}_\mathcal{Z}[D_{KL}(P_{(X|Z,Y|Z)}||P_{X|Z}\otimes P_{Y|Z})].
\end{align}
Note that this equation is an extension of \eqref{eq:mi_kl} to a conditional setting.
The conditional mutual information measures the conditional independence between the relevant variables, \ie, $I(X;Y|Z)=0 \Leftrightarrow X\perp Y|Z$.
Compared to the standard setting, this is particularly important if co-dependence between $X$ and $Y$ changes conditioned on variable $Z$.

We also consider three variables in our problem setting, which include feature representation vector $F$, target variable $Y$, and bias variable $Z$.
As discussed earlier, we concentrate on the mutual information between $F$ and $Z$ conditioned on $Y$ and derive a bias measurement as follows:
\begin{align}
    \text{Cobias} := I(F;Z|Y) &= I(F;Z) - I(F;Z;Y) \\
    &= I(F;Z,Y) - I(F;Y),
    \label{eq:cobias}
\end{align}
where $I(F;Z;Y)$ is an interaction information among three variables and $(Z,Y) \in \mathbb{R}^2$ denotes the joint variables of $Z$ and $Y$.
We estimate the conditional mutual information by computing the difference between two mutual information $I(F;Z,Y)$ and $I(F;Y)$ as expressed in \eqref{eq:cobias}, which is derived from the chain rule of mutual information.

Figure~\ref{fig:diagramB} presents a Venn diagram of the information-theoretic measures for those variables.
Our bias measurement $I(F;Z|Y)$ quantifies the mutual information between feature and bias variable $I(F;Z)$ excluding the amount of the information that also explains the target variable $I(F;Y;Z)$, which corresponds to the region in orange.


\section{Debiasing Frameworks}
\label{sec:method}

This section introduces the proposed two debiasing techniques, which are given by the bias-supervised debiasing regularization and the unsupervised stochastic label noise.

\subsection{Cobias as Debiasing Regularizer}
\label{sec:cobias_sup}
To mitigate the algorithmic bias, we directly incorporate the proposed bias measurement $I(F;Z|Y)$ as an additional loss term to learn our model. 
Then, given the extra supervision about bias attributes $Z$, the final objective function is formulated as
\begin{align}
     &\min_{\theta, \psi} \ell(h(F;\psi), Y) + \beta I(F;Z|Y) \nonumber \\
     &= \min_{\theta, \psi} \ell(h(f(X;\theta);\psi), Y) + \beta I_\phi(f(X;\theta);Z|Y),
\end{align}
where $\ell(\cdot, \cdot)$ is a cross-entropy loss, $\theta$ and $\psi$ denote the parameters of the feature extractor $f(\cdot)$ and the classification layer $h(\cdot)$, respectively, $\phi$ is the parameters of the mutual information estimator network.
The second term in the objective function plays a role as a regularizer to prevent the feature representation from being biased, where $\beta$ is the hyperparameter of its weight.
Note that this regularization term aims to reduce inherent bias within feature representations and has complementary characteristics to the task-specific loss term.
Since the loss does not depend on model architecture or algorithm, it can be easily adopted in other existing debiasing frameworks.
Because the feature representation $F=f(X;\theta)$ is updated during training, the conditional mutual information $I(F;Z|Y)$ also needs to be re-estimated each time, which results in the revision of the objective function as
\begin{align}
     \min_{\theta, \psi} \max_{\phi} \ell(h(f(X;\theta);\psi), Y) + \beta I_\phi(f(X;\theta);Z|Y) \label{eq:minimax}.
\end{align}
Because the objective function is a minimax problem, we alternate to train and update the parameters $(\theta, \psi)$ and $\phi$ in every epoch.

To compute $I(F;Z|Y)$, we should measure the difference between two estimated mutual information values, $I_{\phi_1}(F;Z,Y)$ and $I_{\phi_2}(F;Y)$ as shown in \eqref{eq:cobias}.
However, minimizing the difference of the two values obtained from the two different estimators parameterized by $\phi_1$ and $\phi_2$ may hamper training stability.
We sidestep this issue by minimizing a surrogate mutual information, $I(F,Y;Z)$, with respect to $\theta$, instead of $I(F;Z|Y)$.
Because $I(f(X;\theta),Y;Z) = I(f(X;\theta);Z|Y) + I(Y;Z)$ and $I(Y;Z)$ is a constant independent of $\theta$, minimizing $I(f(X;\theta);Z|Y)$ with respect to $\theta$ is equivalent to minimizing $I(f(X;\theta),Y;Z)$.
This enables us to train the network based on \eqref{eq:minimax} using a single estimator network and consequently facilitates stable training.

\subsection{Stochastic Label Noise}

We now introduce a simple yet effective unsupervised debiasing approach exploiting synthetic label noise.
Stochastic label noise perturbs the target label from its true class to any other class with a noise rate $\rho$ on each mini-batch independently.
Let $Y$ and $\widetilde{Y}$ be the true and noisy target labels, respectively.
Then, $p(\widetilde{Y} = k|Y=y) = \frac{\rho}{K-1}$ for all $k \neq y$ and $1-\rho$ for $k = y$, where $K$ is the number of classes.
The objective function is given by
\begin{align}
\min_{\theta, \psi}\ell(h(f(X;\theta);\psi), \widetilde{Y}).
\end{align}

Stochastic label noise, even without any explicit supervision of bias information, mitigates the algorithmic bias while minimizing the performance degradation in the classification for target variable.
Intuitively, this is because spurious correlation is more susceptible to label noise than true causation.
Label noise gives an implicit ensemble effect~\cite{xie2016disturblabel}, which helps to find invariant properties between a feature representation and its target label. 
Because spurious correlation does not appear to be stable properties~\cite{IRM, woodward2005making}, label noise helps to absorb spurious correlation and mitigate algorithmic bias.

From the perspective of information theory, it is natural that injecting label noise reduces the mutual information between target variable $Y$ and bias variable $Z$, \ie, $I(Z;\widetilde{Y}) < I(Z;Y)$. 
In addition, by adding stochastic label noise, the reduction of mutual information between $Y$ and $Z$ turns out to be more significant than that of $Y$'s entropy.
According to our experiment, the ratio $R := {I(Z;\widetilde{Y})} / {I(Y;\widetilde{Y})}$ is 0.187 in the absence of label noise with $Y=$~``\textit{hair color}" and $Z=$~``\textit{gender}" in the CelebA dataset.
However, it decreases when we add several different levels of label noise, and the tendency is consistent in other $(Y, Z)$ pairs if they are correlated.
This observation implies that injecting label noise is helpful for alleviating spurious correlation without affecting classification performance.
Note that this label perturbation method is orthogonal to existing debiasing frameworks and can be applied to them with no modification.


\section{Experiments}

\subsection{Experimental Setup}
\label{sec:exp_setup}

\paragraph{Datasets}
We conduct experiments on the three standard benchmarks: CelebA~\cite{CelebA}, Waterbirds~\cite{GroupDRO}, and FairFace~\cite{Fairface}.
CelebA is a large-scale face dataset composed of 202,599 celebrity images with 40 attributes.
This dataset is available for non-commercial research purposes.
We follow the original train-val-test split~\cite{CelebA} throughout the experiments.
Waterbirds~\cite{GroupDRO} is a synthesized dataset with 4,795 training examples, which are created by combining bird images in the CUB dataset~\cite{wah2011caltech} and background images from the Places dataset~\cite{zhou2017places}.
Following the setup in \cite{GroupDRO}, each image has two attributes; one is the type of bird, $\{$waterbird, landbird$\}$ and the other is the background place, $\{$water, land$\}$.
FairFace~\cite{Fairface} is a recently proposed face image dataset containing 108,501 images, which are collected from the YFCC-100M Flickr dataset~\cite{thomee2016yfcc100m}.
The dataset has seven race groups\footnote{\textit{White}, \textit{Black}, \textit{Indian}, \textit{East Asian}, \textit{Southeast} \textit{Asian}, \textit{Middle Eastern}, and \textit{Latino}.}, nine age groups, and two gender groups.

\paragraph{Implementation details}
We use the ResNet-18~\cite{he2016deep} pretrained on ImageNet~\cite{deng2009imagenet} as our backbone network for all experiments.
We train our models using the stochastic gradient descent method with the Adam optimizer for 50 epoch.
The learning rate is $1 \times 10^{-4}$, and the batch size is 256.
We set a weight decay to $1 \times 10^{-2}$ for Waterbirds and  $1 \times 10^{-4}$ for the other two datasets.
The weight of the bias regularizer $\beta$ is fixed to 5.
The noise rate $\rho$ is set to 0.1 for Waterbirds and 0.2 for the rest.
For all methods, we leverage a simple resampling technique based on the class size to alleviate the class imbalance issue.
Our algorithms are implemented in the Pytorch~\cite{paszke2019pytorch} framework and all experiments are conducted on a single unit of NVIDIA Titan XP GPU.

\paragraph{Evaluation metrics}
In addition to average accuracy, we evaluate all the compared algorithms with three main metrics, Cobias, unbiased accuracy, and worst-group accuracy, to provide a comprehensive view of the algorithmic bias.
We also adopt other fairness metrics such as bias amplification (BA)~\cite{zhao2017men}, equalized opportunity difference (EO), and disparate impact (DI)~\cite{hardt2016equality}, where low values are preferred for the extra metrics.

\subsection{Results}

\begin{table*}[t]
\begin{center}
\caption{Experimental results of our debiasing frameworks on the test split of the CelebA dataset.
We set \textit{gender} attribute to the bias variable.
}
\label{tab:celeba}
 \scalebox{0.85}{
\setlength\tabcolsep{5.0pt} \hspace{-0.25cm}
\begin{tabular}{lc|c|ccccc:c}
\toprule
& Target & Cobias & Unbiased Acc. &Worst-group Acc.& ~~~~~BA~~~~~ & ~~~~~EO~~~~~ & ~~~~~DI~~~~~ &Average Acc. \\
\hline
ERM & Hair Color 				& 0.435 & 84.1 & 53.2 & ~0.014 & 0.48 & 0.113 & \bf{95.1} \\
ERM + label noise & Hair Color		& 0.286 & 87.8 & 66.9 & \bf{-0.002} & 0.35 & \bf{0.036} & 93.8\\
ERM + bias regularizer& Hair Color	&\bf{0.111} & \bf{88.1} & \bf{67.4} & ~0.001 & \bf{0.31} & 0.039 & 94.5 \\
\hdashline
Group DRO~\cite{GroupDRO} & Hair Color 	& 0.409 & 89.4 & 77.6 & ~0.005 & 0.17 & 0.044 & \bf{94.2}\\
Group DRO + label noise & Hair Color 		& 0.270 & {90.6} & {84.0} & \bf{-0.004} & \bf{0.07} & 0.030 & 93.5\\
Group DRO + bias regularizer & Hair Color 	& \bf{0.045} & \bf{90.7}& \bf{85.4} & ~0.002 & 0.09 & \bf{0.018} & 94.0 \\
\hline
ERM & Pale Skin 				& 0.387 & 82.7 &59.6 & ~0.012 & 0.38 & 0.273 & \bf{95.5} \\
ERM + label noise & Pale Skin		& 0.192 & \bf{87.6} & \bf{73.8} & \bf{~0.002} & \bf{0.19} & \bf{0.049} & 94.7 \\
ERM + bias regularizer&Pale Skin	& \bf{0.066} & {86.1} & {68.7} & ~0.006 & 0.28 & 0.153 &  95.2 \\
\hdashline
Group DRO~\cite{GroupDRO} & Pale Skin 	& 0.346 & 88.8 & 84.6 & ~0.005 & 0.11 & 0.109 & 93.8 \\
Group DRO + label noise & Pale Skin 		& 0.189 & \bf{90.3} & \bf{86.4} & \bf{~0.001} & \bf{0.06} & \bf{0.002} & \bf{94.2} \\
Group DRO + bias regularizer & Pale Skin 	& \bf{0.114} & 89.1 & 85.5 & ~0.003 & 0.10 & 0.063& 93.9 \\
\hline
ERM & Smiling 					& 0.126 & 92.1 & 88.6 & ~0.002 & \bf{0.01} & \bf{0.008} & 92.5\\
ERM + label noise & Smiling 		& 0.017 & \bf{92.4} & 88.7 & \bf{-0.015} & 0.10 & 0.057 & \bf{92.8} \\
ERM + bias regularizer & Smiling 	& \bf{0.011} & \bf{92.4} & \bf{89.0} & -0.011 & 0.08 & 0.038 &  \bf{92.8} \\
\hdashline
Group DRO~\cite{GroupDRO} & Smiling 	& 0.130 & 92.1 & 89.6 & ~0.003 & 0.02 & 0.012 & 92.2 \\
Group DRO + label noise & Smiling 		& \bf{0.019} & 92.5 & \bf{90.1} & -0.003 & \bf{0.01} & \bf{0.011} & 92.6 \\
Group DRO + bias regularizer & Smiling 	& 0.024 & \bf{92.7} & 89.9 & \bf{-0.005} & \bf{0.01} & 0.019 & \bf{92.8} \\
\bottomrule
\end{tabular}
 }
\end{center}
\end{table*}

\paragraph{CelebA}
Table~\ref{tab:celeba} presents the experiment results on the test split of the CelebA dataset. 
Among all attributes, we choose \textit{blond hair}, \textit{pale skin}, and \textit{smiling} as target variables while \textit{gender} is used as the bias variable, setting up three different classification tasks.
We employ two baseline algorithms, including empirical risk minimization (ERM) and group distributionally robust optimization (Group DRO)~\cite{GroupDRO}, to which the proposed two debiasing methods are applied.
As shown in Table~\ref{tab:celeba}, incorporating the bias regularization loss significantly reduces Cobias and other fairness scores, while preserving the average classification accuracy.
This implies that the proposed bias measurement is meaningful for identifying the algorithmic bias.
We also observe that applying stochastic label noise is effective to mitigate the algorithmic bias even without extra supervision, which validates our claim that spurious correlation is more sensitive to label noise than true causation.
Note that Group DRO yields high bias scores in spite of its high group-fairness accuracy, which implies that it may adjust the parameters for bias-related features in the classification layer instead of learning debiased representations.

\begin{table}[t]
\begin{center}
\caption{Experimental results on the test split of the Waterbirds dataset, where bird type and background place are set to target and bias variables, respectively.
}
\label{tab:waterbirds}
 \scalebox{0.85}{
\setlength\tabcolsep{4pt} \hspace{-0.27cm}
\begin{tabular}{l|c|cc:c}
\toprule
& Cobias & Unbiased & Worst & Average \\
\hline
ERM & 0.482 & 81.4 & 52.3 & 83.6 \\
ERM + label noise & 0.413 & 83.0 & 58.2 & 83.1 \\
ERM + bias regularizer & \bf{0.107} & \bf{84.3} & \bf{72.2} & \bf{88.4} \\
\hdashline
Group DRO & 0.376 & 86.3 & 72.4 & 85.7 \\
Group DRO + label noise &  0.348 & \bf{86.7} & 70.6 & 84.7 \\
Group DRO + bias regularizer & \bf{0.098} & \bf{86.7} & \bf{76.9} & \bf{87.4} \\
\bottomrule
\end{tabular}
 }
\end{center}
 \vspace{-0.1cm}
\end{table}

\begin{table}[t]
\begin{center}
\caption{Experimental results on the test split of the Fairface dataset, where \textit{age} and \textit{race} are set to target and bias variables, respectively.
}
 \vspace{-0.2cm}
\label{tab:fairface}
 \scalebox{0.85}{
\setlength\tabcolsep{4pt} \hspace{-0.27cm}
\begin{tabular}{l|c|cc:c}
\toprule
& Cobias & Unbiased & Worst & Average \\
\hline
ERM & 0.019 & 47.6 & 16.9 & 52.1 \\
ERM + label noise & 0.017 & 48.7 & \bf{18.6} & 53.2 \\
ERM + bias regularizer & \bf{0.004} & \bf{49.6} & 18.5 & \bf{53.4} \\
\hdashline
Group DRO & 0.015 & 48.4 & 18.8 & 51.5 \\
Group DRO + label noise & 0.004 & 49.2 & 16.8 & 52.5 \\
Group DRO + bias regularizer & \bf{0.002} &  \bf{49.4} & \bf{19.3} & \bf{53.6} \\
\bottomrule
\end{tabular}
 }
\end{center}
 \vspace{-0.1cm}
\end{table}

\paragraph{Waterbirds}

The results on the Waterbirds dataset are presented in Table~\ref{tab:waterbirds}, where the bird type and the background place are set to target and bias variables, respectively.
According to our experiments, the proposed debiasing frameworks consistently produce promising results in this small-scale synthesized dataset.
Compared to the CelebA dataset, injecting label noise is less effective to reduce Cobias in this dataset, which is partly due to its small size.

\paragraph{FairFace}

We also evaluate our frameworks on the FairFace dataset and demonstrate the results in Table~\ref{tab:fairface}, where the target is \textit{age} and the bias is \textit{race}.
Our frameworks still improves both Cobias and group-fairness accuracy when combined with both ERM and Group DRO algorithms.
Note that the Cobias scores are particularly small because the FairFace dataset is constructed with an emphasis of balanced race composition, as stated in~\cite{Fairface}.

\subsection{Analysis}

\paragraph{Domain generalization scenario}

\begin{table*}[t]
\begin{center}
\caption{Experimental results in domain generalization setting on the CelebA dataset, where the bias variable is fixed to \textit{gender} for all settings.
}
\vspace{-0.1cm}
\label{tab:dg}
 \scalebox{0.85}{
\setlength\tabcolsep{5.3pt} \hspace{-0.25cm}
\begin{tabular}{lccc|c|cc:c}
\toprule
& Target & Domain (train) & Domain (test) & ~Cobias~ & Unbiased Acc. & Worst-group Acc.  & Average Acc. \\
\hline
ERM & Hair Color & Old & Young & 0.432 & 81.5 & 43.5 & 93.8 \\
ERM + label noise & Hair Color & Old & Young  &0.367 & 88.1 & 68.3 & \bf{93.9}\\
ERM + bias regularizer& Hair Color & Old & Young & \bf{0.272} & \bf{90.4} & \bf{79.7} & 93.0 \\
\hdashline
Group DRO~\cite{GroupDRO} & Hair Color & Old & Young  & 0.387 & 86.2 & 68.1 & \bf{94.2} \\
Group DRO + label noise & Hair Color& Old & Young  & 0.171 & 88.6 & 77.4 & 93.9  \\
Group DRO + bias regularizer& Hair Color & Old & Young & \bf{0.061} & \bf{89.5} & \bf{78.6} & 94.1  \\
\hline
ERM & Hair Color & Slim & Chubby & 0.234 & 75.6 & 29.5 & \bf{98.1} \\
ERM + label noise & Hair Color & Slim & Chubby & 0.107 & {82.7} & \bf{54.5} & 97.9\\
ERM + bias regularizer& Hair Color & Slim & Chubby  & \bf{0.073} & \bf{84.5} & 54.2 & 97.9\\
\hdashline
Group DRO~\cite{GroupDRO} &  Hair Color&Slim & Chubby & 0.192 & 84.5 & 67.6 & 96.5\\
Group DRO + label noise &  Hair Color  & Slim & Chubby & 0.062 & \bf{87.2} & 64.2  & 96.8\\
Group DRO + bias regularizer&  Hair Color & Slim & Chubby & \bf{0.029} & 87.1 & \bf{69.7} & \bf{96.9} \\
\bottomrule
\end{tabular}
 }
\end{center}
\end{table*}

We extend our task to domain generalization scenario, where training and test sets belong to different domains.
This scenario is realistic because domain and distribution shifts may occur simultaneously.
To build this setup, we introduce another variable, other than for target or bias attributes, which is for domain attribute.
The examples in the training and test datasets are not supposed to have the same value for the domain variable. 
For example, if we have a domain attribute, \textit{Young}, then training examples may be composed of the images with the old (\textit{Young = false}) while test dataset may contain the images of young people only (\textit{Young = true}).
Since the domain attribute, \textit{Young} in this case, is correlated to both the target and bias variables, it affects the group distributions in the training and test sets significantly and they are not consistent no longer.
Table~\ref{tab:dg} presents the domain generalization results with two domain attributes, \textit{Young} and \textit{Chubby}, where our frameworks outperform the baselines consistently in the presence of domain shift.
We also observe that ERM with the bias regularizer often gives better results than the original Group DRO method.
This results validate that our frameworks are particularly effective when domain and distribution shifts exist at the same time.

\begin{table}[t]
    \centering
    \captionof{table}{
    Ablative results for the weight parameter $\beta$ of the bias regularization loss on the CelebA dataset.
    }
        \vspace{-0.1cm}
    \label{tab:abl_regularizer}
     \scalebox{0.85}{
\setlength\tabcolsep{8.5pt} \hspace{-0.25cm}
\begin{tabular}{c|c|cc}
\toprule
Bias weight & ~~Cobias~~ & Unbiased Acc. & Worst-group Acc. \\
\hline
0 & 0.435 & 84.1 & 53.2 \\
1 & 0.202 & 84.7 & 58.3  \\
2 & 0.183 & 87.2 & 62.1  \\
5 & {0.111} & \bf{88.1} & \bf{67.4}\\
10 & \bf{0.046} & {87.3} & {65.6}\\
\bottomrule
\end{tabular}
}
\end{table}

\begin{table}[t]
    \centering
    \captionof{table}{
    Ablative results for the noise rate $\rho$ in the stochastic label noise on the CelebA dataset.
    }
    \vspace{-0.1cm}
    \label{tab:abl_noise}
     \scalebox{0.85}{
\setlength\tabcolsep{9pt} \hspace{-0.25cm}
\begin{tabular}{c|c|cc}
\toprule
Noise rate& ~~Cobias~~ & Unbiased Acc. & Worst-group Acc.  \\
\hline
0.0 & 0.435 & 84.1 & 53.2 \\
0.1 & 0.362 & {87.3} & {65.7}  \\
0.2 & 0.286 & \bf{87.8} & \bf{66.9} \\
0.4 & \bf{0.209} & 86.1 & 60.6\\
\bottomrule
\end{tabular}
}
\end{table}

\paragraph{Ablation study}

To validate the effectiveness of the bias regularization loss and the stochastic label noise scheme, we perform the ablation study on the weight parameter $\beta$ and the noise rate $\rho$.
Table~\ref{tab:abl_regularizer} and~\ref{tab:abl_noise} show the ablative results on the CelebA dataset with \textit{hair color} and \textit{gender} for target and bias variables, respectively.
The results show that increasing the weight of the bias regularization loss and the noise rate improves both Cobias and group-fairness accuracy within sufficiently wide ranges.

\paragraph{Feature visualization}

\begin{figure*}[t]
\centering
\hspace*{0.3cm}
    \begin{subfigure}[m]{0.23\linewidth}
    	\includegraphics[width=\linewidth]{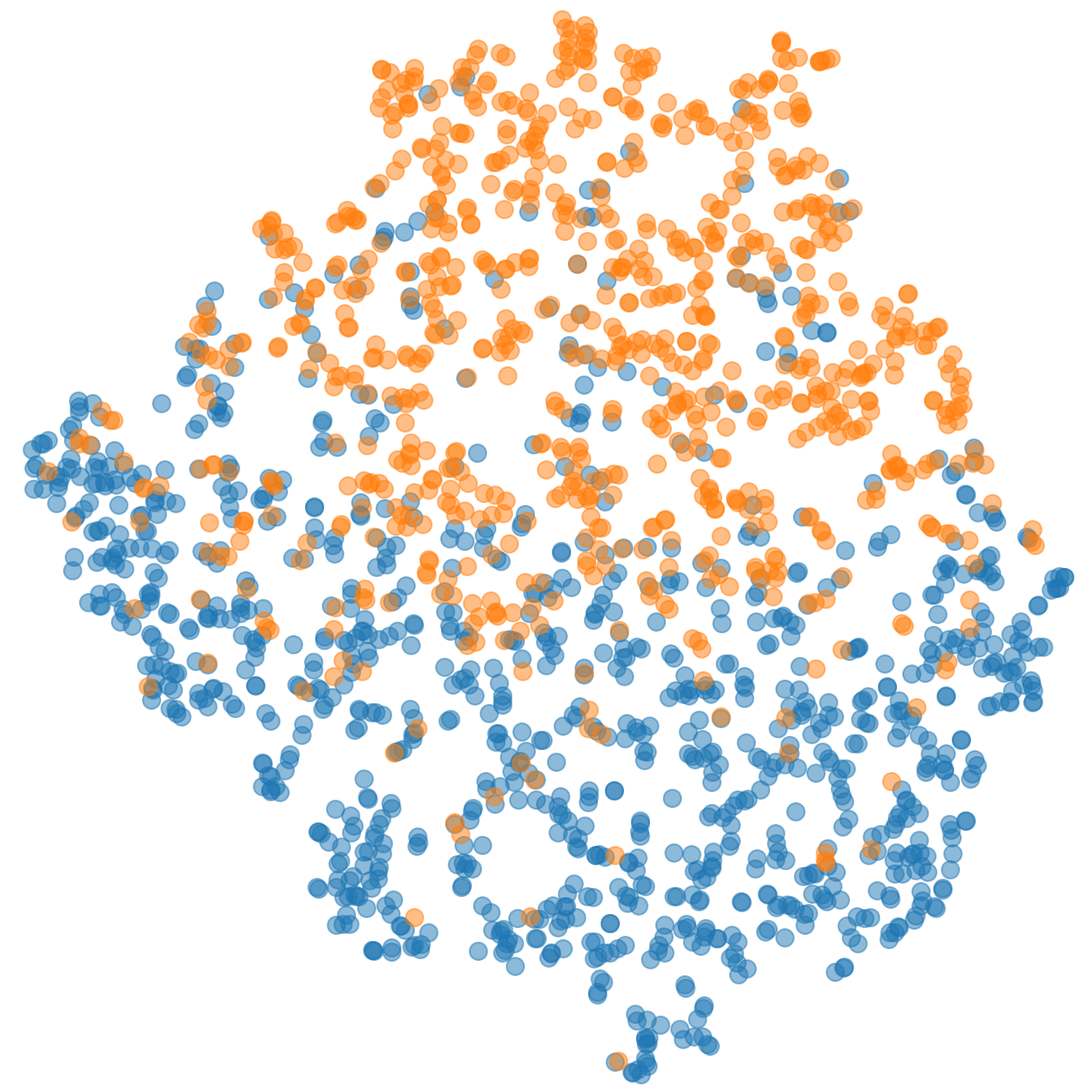}
	\subcaption{ERM}
	\label{fig:tsne_erm}
	\end{subfigure} 
	\hspace{0.5cm}
    \begin{subfigure}[m]{0.23\linewidth}
    	\includegraphics[width=\linewidth]{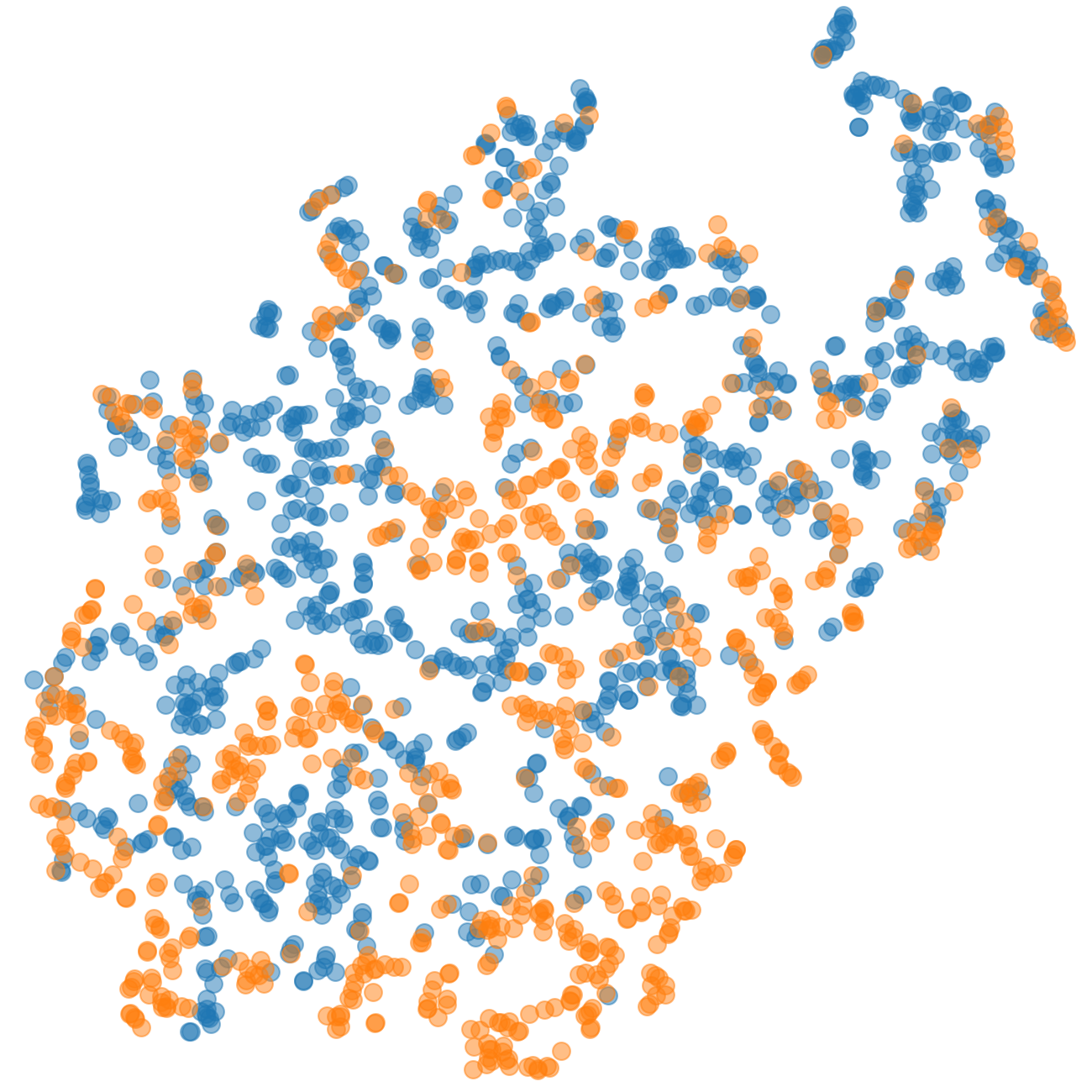}
	\subcaption{ERM + label noise}
	\label{fig:tsne_noise}
    \end{subfigure} 
    \hspace{0.5cm}
    \begin{subfigure}[m]{0.23\linewidth}
    	\includegraphics[width=\linewidth]{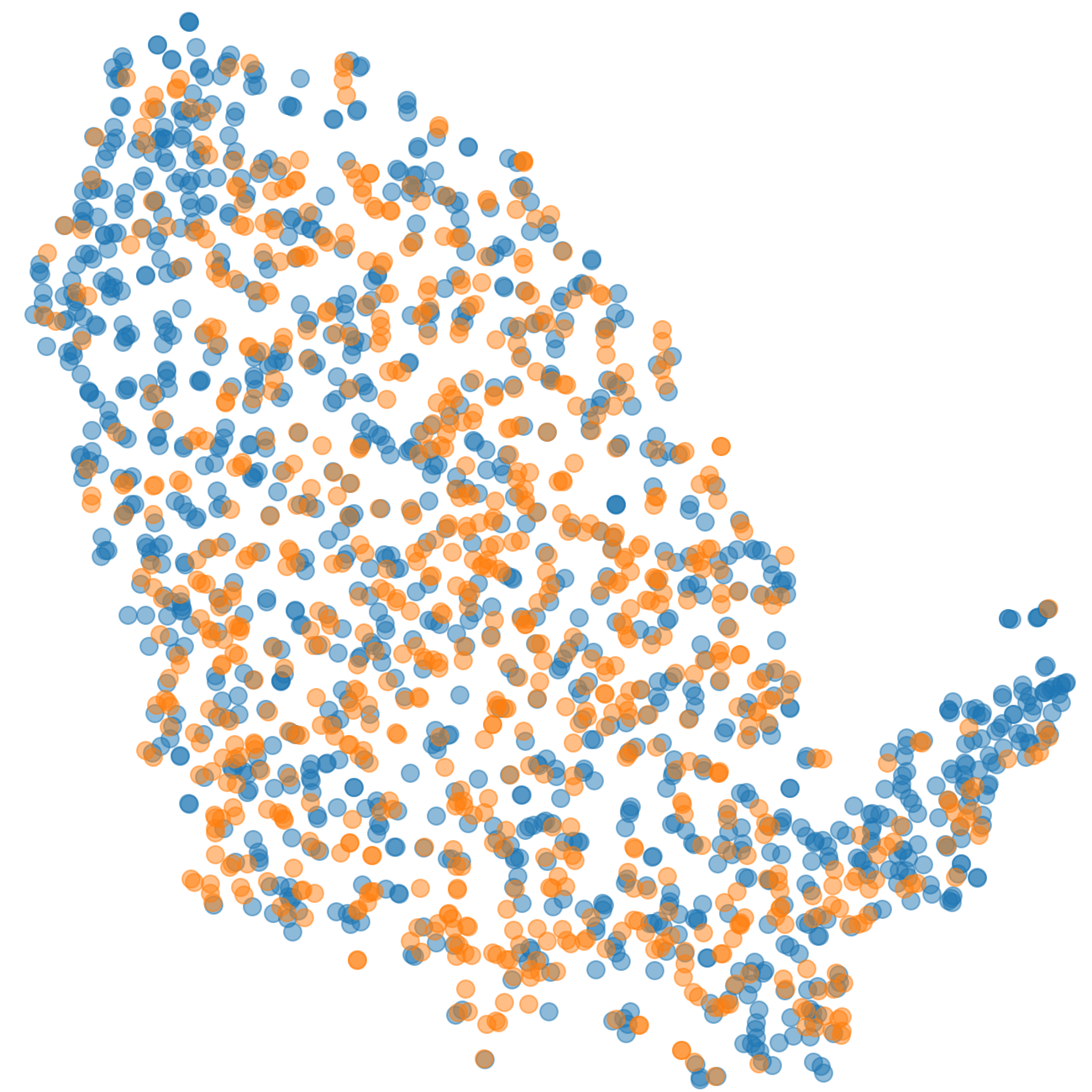}
	\subcaption{ERM + bias regularizer}
	\label{fig:tsne_bias}
    \end{subfigure} 
    \caption{The t-SNE plots of feature representations from the ResNet-18 models trained with ERM and our debiasing frameworks on the CelebA dataset using the \textit{hair color} classification.
    We visualize the distribution of samples who have the same target value (blond hair).
    Blue and orange colors denote different gender values (female and male, respectively).
    }
    \label{fig:tsne}
\end{figure*}

Figure~\ref{fig:tsne} illustrates the t-SNE visualizations~\cite{van2008visualizing} of the feature embeddings of the samples in the CelebA test split given by \textit{hair color} classification.
For simplicity, we visualize only the examples with blond hair.
Blue and orange colors denote female and male, values of the \textit{gender} bias variable, respectively.
Figure~\ref{fig:tsne_erm}, even without using the bias information (\textit{gender}) for training, shows that the examples drawn from a bias group are distinguishable from those from the other bias group.
Compared to vanilla ERM, as illustrated in Figure~\ref{fig:tsne_noise} and~\ref{fig:tsne_bias}, ERMs with bias regularization and stochastic label noise successfully make the examples with different bias attributes confused on the feature embedding space, which is desirable for learning debiased representations.

\paragraph{Comparison to logit-based bias regularizer}

\begin{table*}[t]
\begin{center}
\caption{Comparison to the logit-based bias regularization loss on the CelebA dataset with \textit{gender} bias.
}
\label{tab:logit_based}
 \scalebox{0.85}{
\setlength\tabcolsep{12pt} \hspace{-0.25cm}
\begin{tabular}{lc|c|cc:c}
\toprule
Bias regularizer type& Target & Cobias & Unbiased Acc. & Worst-group Acc. & Average Acc. \\
\hline
logit-based & Hair Color & 0.219 & 84.9 & 62.7 & \bf{94.6} \\
feature-based (Ours)& Hair Color &\bf{0.111} & \bf{88.1} & \bf{67.4} & 94.5 \\ 
\hline
logit-based & Pale Skin & 0.182 & 84.0 & 60.8 & \bf{95.7} \\
feature-based (Ours)& Pale Skin & \bf{0.066} & \bf{86.1} & \bf{68.7} & 95.2 \\
\bottomrule
\end{tabular}
 }
\end{center}
\end{table*}
To analyze the effectiveness of our feature-level bias measurement, we compare the feature-based bias regularizer $I(F;Z|Y)$ in~\eqref{eq:minimax} with logit-based bias regularizer $I(\hat{Y};Z|Y)$, where $\hat{Y} = h(F) \in \mathcal{Y}$ is a predicted target variable from a classifier based on a fully connected layer, $h(\cdot)$.
Table~\ref{tab:logit_based} presents the results from the two regularizers, where the feature-based approach outperforms its logit-based counterpart significantly in terms of both Cobias and group-fairness accuracy.
This means that exploiting feature representation would be more effective to identify and mitigate the algorithmic bias than logits.


\section{Related Work} 
\label{sec:related_work}
Facial recognition datasets often contain inevitable biases due to their insufficiently controlled data collection process, and consequently, recognition algorithms tend to inherit and even amplify the dataset bias.
To address this issue, numerous debiasing frameworks have been proposed for identifying and mitigating the potential risks posed by dataset or algorithmic bias.
These frameworks can be categorized in pre-processing~\cite{li2019repair, sagawa2020investigation, kamiran2012data}, in-processing~\cite{GroupDRO, sohoni2020no, wang2019iccv, zhang2018mitigating, gong2020jointly, seo2021unsupervised, ragonesi2021learning, wang2020towards, guo2020learning}, and post-processing~\cite{hardt2016equality, zhao2017men} ones.
Pre-processing techniques transform the data distribution to keep the training data balanced across groups, where dataset resampling or reweighting methods~\cite{li2019repair, sagawa2020investigation, kamiran2012data} are usually exploited to balance the distribution by under-sampling the majority or over-sampling the minority classes.
Debiasing through in-processing aims to build algorithms that can learn fair representation, by taking advantage of adversarial training~\cite{wang2019iccv, zhang2018mitigating}, representation disentanglement~\cite{gong2020jointly, ragonesi2021learning}, or robust optimization~\cite{GroupDRO, sohoni2020no, seo2021unsupervised}.
Post-processing methods modify the predicted outputs to meet fairness criterion, mainly by calibrating the outputs~\cite{hardt2016equality, zhao2017men}.
Our debiasing frameworks belong to in-processing methods, which adopts a bias regularization loss or injecting label noise during training to reduce the algorithmic bias.
The bias regularization loss is somewhat similar to \cite{ragonesi2021learning}, but the major difference is that we introduce \emph{conditional} mutual information from the structural causal model. 
The use of unconditional mutual information as a bias regularizer makes it difficult to achieve our goal, learning the relationship between features and target variables.

Mutual information is an information-theoretic quantity to measure the relationship between two variables.
Because it can capture non-linear dependencies between the variables, the mutual information is widely used in various areas, including unsupervised representation learning~\cite{comon1994independent, tishby2000information, oord2018representation, hjelm2018learning, sun2019infograph}, generative models~\cite{chen2016infogan, qian2019enhancing}, reinforcement learning~\cite{oord2018representation}, and fair supervised learning~\cite{kamishima2012fairness, fukuchi2015prediction, ragonesi2021learning}.
However, the exact computation of mutual information between continuous variables is basically not tractable.
There exists some non-parametric estimators based on kernel density estimation~\cite{kwak2002input, suzuki2008approximating} to deal with continuous variables, but those are not scalable with high-dimensional or large-scale data.
To overcome this limitation, recent works on mutual information estimation focus on training neural network to represent its variational lower bound~\cite{belghazi2018mutual, lin2019data, poole2019variational}.
Our bias measurement takes also advantage of the variational lower bound~\cite{donsker1975asymptotic, belghazi2018mutual} of conditional mutual information.


\section{Conclusion} 
\label{sec:conclusion}

We proposed an information-theoretic bias measurement which can identify the feature-level algorithmic bias from the causal view of spurious correlation.
Based on the new measurement approach, we presented two types of debiasing frameworks with bias regularizer or label noise, each of which can be utilized with or without explicit knowledge of bias information, respectively.
We demonstrated the effectiveness and versatility of proposed frameworks on multiple standard benchmarks.
We also conducted a detailed analysis of our measurement and frameworks via extensive ablation studies with more realistic scenarios.

\subsection*{Acknowledgement}

This work was partly supported by Samsung Advanced Institute of Technology, the Bio \& Medical Technology Development Program of the National Research Foundation (NRF) funded by the Korea government (MSIT) [No. 2021M3A9E4080782], and  Institute of Information \& communications Technology Planning \& Evaluation (IITP) grant funded by the Korean government (MSIT) [No.2021-0-01343, Artificial Intelligence Graduate School Program (Seoul National University)].

{\small
\bibliographystyle{ieee_fullname}
\bibliography{aaai22}

\begin{thebibliography}{50}
\providecommand{\natexlab}[1]{#1}

\bibitem[{Arjovsky et~al.(2019)Arjovsky, Bottou, Gulrajani, and
  Lopez-Paz}]{IRM}
Arjovsky, M.; Bottou, L.; Gulrajani, I.; and Lopez-Paz, D. 2019.
\newblock Invariant risk minimization.
\newblock \emph{arXiv preprint arXiv:1907.02893}.

\bibitem[{Belghazi et~al.(2018)Belghazi, Baratin, Rajeshwar, Ozair, Bengio,
  Courville, and Hjelm}]{belghazi2018mutual}
Belghazi, M.~I.; Baratin, A.; Rajeshwar, S.; Ozair, S.; Bengio, Y.; Courville,
  A.; and Hjelm, D. 2018.
\newblock Mutual information neural estimation.
\newblock In \emph{ICML}.

\bibitem[{Buolamwini and Gebru(2018)}]{buolamwini2018gender}
Buolamwini, J.; and Gebru, T. 2018.
\newblock Gender shades: Intersectional accuracy disparities in commercial
  gender classification.
\newblock In \emph{ACM FAccT}.

\bibitem[{Cadene et~al.(2019)Cadene, Dancette, Cord, Parikh et~al.}]{RUBi}
Cadene, R.; Dancette, C.; Cord, M.; Parikh, D.; et~al. 2019.
\newblock Rubi: Reducing unimodal biases for visual question answering.
\newblock In \emph{NeurIPS}.

\bibitem[{Calders, Kamiran, and Pechenizkiy(2009)}]{calders2009building}
Calders, T.; Kamiran, F.; and Pechenizkiy, M. 2009.
\newblock Building classifiers with independency constraints.
\newblock In \emph{ICDM}.

\bibitem[{Chen et~al.(2016)Chen, Duan, Houthooft, Schulman, Sutskever, and
  Abbeel}]{chen2016infogan}
Chen, X.; Duan, Y.; Houthooft, R.; Schulman, J.; Sutskever, I.; and Abbeel, P.
  2016.
\newblock Infogan: Interpretable representation learning by information
  maximizing generative adversarial nets.
\newblock In \emph{NIPS}.

\bibitem[{Comon(1994)}]{comon1994independent}
Comon, P. 1994.
\newblock Independent component analysis, a new concept?
\newblock \emph{Signal processing}, 36(3): 287--314.

\bibitem[{Deng et~al.(2009)Deng, Dong, Socher, Li, Li, and
  Fei-Fei}]{deng2009imagenet}
Deng, J.; Dong, W.; Socher, R.; Li, L.-J.; Li, K.; and Fei-Fei, L. 2009.
\newblock Imagenet: A large-scale hierarchical image database.
\newblock In \emph{CVPR}.

\bibitem[{Donsker and Varadhan(1975)}]{donsker1975asymptotic}
Donsker, M.~D.; and Varadhan, S.~S. 1975.
\newblock Asymptotic evaluation of certain Markov process expectations for
  large time, I.
\newblock \emph{Communications on Pure and Applied Mathematics}, 28(1): 1--47.

\bibitem[{Dwork et~al.(2012)Dwork, Hardt, Pitassi, Reingold, and
  Zemel}]{dwork2012fairness}
Dwork, C.; Hardt, M.; Pitassi, T.; Reingold, O.; and Zemel, R. 2012.
\newblock Fairness through awareness.
\newblock In \emph{ITCS}.

\bibitem[{Fukuchi, Kamishima, and Sakuma(2015)}]{fukuchi2015prediction}
Fukuchi, K.; Kamishima, T.; and Sakuma, J. 2015.
\newblock Prediction with model-based neutrality.
\newblock \emph{IEICE TRANSACTIONS on Information and Systems}, 98(8):
  1503--1516.

\bibitem[{Gong, Liu, and Jain(2020)}]{gong2020jointly}
Gong, S.; Liu, X.; and Jain, A.~K. 2020.
\newblock Jointly de-biasing face recognition and demographic attribute
  estimation.
\newblock In \emph{ECCV}.

\bibitem[{Guo et~al.(2020)Guo, Zhu, Zhao, Cao, Lei, and Li}]{guo2020learning}
Guo, J.; Zhu, X.; Zhao, C.; Cao, D.; Lei, Z.; and Li, S.~Z. 2020.
\newblock Learning meta face recognition in unseen domains.
\newblock In \emph{CVPR}.

\bibitem[{Hardt, Price, and Srebro(2016)}]{hardt2016equality}
Hardt, M.; Price, E.; and Srebro, N. 2016.
\newblock Equality of opportunity in supervised learning.
\newblock In \emph{NIPS}.

\bibitem[{He et~al.(2016)He, Zhang, Ren, and Sun}]{he2016deep}
He, K.; Zhang, X.; Ren, S.; and Sun, J. 2016.
\newblock {Deep Residual Learning for Image Recognition}.
\newblock In \emph{CVPR}.

\bibitem[{Hjelm et~al.(2019)Hjelm, Fedorov, Lavoie-Marchildon, Grewal, Bachman,
  Trischler, and Bengio}]{hjelm2018learning}
Hjelm, R.~D.; Fedorov, A.; Lavoie-Marchildon, S.; Grewal, K.; Bachman, P.;
  Trischler, A.; and Bengio, Y. 2019.
\newblock Learning deep representations by mutual information estimation and
  maximization.
\newblock In \emph{ICLR}.

\bibitem[{Kamiran and Calders(2012)}]{kamiran2012data}
Kamiran, F.; and Calders, T. 2012.
\newblock Data preprocessing techniques for classification without
  discrimination.
\newblock \emph{Knowledge and Information Systems}, 33(1): 1--33.

\bibitem[{Kamishima et~al.(2012)Kamishima, Akaho, Asoh, and
  Sakuma}]{kamishima2012fairness}
Kamishima, T.; Akaho, S.; Asoh, H.; and Sakuma, J. 2012.
\newblock Fairness-aware classifier with prejudice remover regularizer.
\newblock In \emph{ECML PKDD}.

\bibitem[{K{\"a}rkk{\"a}inen and Joo(2021)}]{Fairface}
K{\"a}rkk{\"a}inen, K.; and Joo, J. 2021.
\newblock Fairface: Face attribute dataset for balanced race, gender, and age.
\newblock In \emph{WACV}.

\bibitem[{Kim(2016)}]{kim2016data}
Kim, P.~T. 2016.
\newblock Data-driven discrimination at work.
\newblock \emph{Wm. \& Mary L. Rev.}, 58: 857.

\bibitem[{Koh et~al.(2021)Koh, Sagawa, Marklund, Xie, Zhang, Balsubramani, Hu,
  Yasunaga, Phillips, Gao, Lee, David, Stavness, Guo, Earnshaw, Haque, Beery,
  Leskovec, Kundaje, Pierson, Levine, Finn, and Liang}]{wilds2021}
Koh, P.~W.; Sagawa, S.; Marklund, H.; Xie, S.~M.; Zhang, M.; Balsubramani, A.;
  Hu, W.; Yasunaga, M.; Phillips, R.~L.; Gao, I.; Lee, T.; David, E.; Stavness,
  I.; Guo, W.; Earnshaw, B.~A.; Haque, I.~S.; Beery, S.; Leskovec, J.; Kundaje,
  A.; Pierson, E.; Levine, S.; Finn, C.; and Liang, P. 2021.
\newblock WILDS: A Benchmark of in-the-Wild Distribution Shifts.
\newblock \emph{arXiv}.

\bibitem[{Kwak and Choi(2002)}]{kwak2002input}
Kwak, N.; and Choi, C.-H. 2002.
\newblock Input feature selection by mutual information based on Parzen window.
\newblock \emph{TPAMI}, 24(12): 1667--1671.

\bibitem[{Li and Vasconcelos(2019)}]{li2019repair}
Li, Y.; and Vasconcelos, N. 2019.
\newblock Repair: Removing representation bias by dataset resampling.
\newblock In \emph{CVPR}.

\bibitem[{Lin et~al.(2019)Lin, Sur, Nastase, Divakaran, Hasson, and
  Amer}]{lin2019data}
Lin, X.; Sur, I.; Nastase, S.~A.; Divakaran, A.; Hasson, U.; and Amer, M.~R.
  2019.
\newblock Data-efficient mutual information neural estimator.
\newblock \emph{arXiv preprint arXiv:1905.03319}.

\bibitem[{Liu et~al.(2015)Liu, Luo, Wang, and Tang}]{CelebA}
Liu, Z.; Luo, P.; Wang, X.; and Tang, X. 2015.
\newblock Deep learning face attributes in the wild.
\newblock In \emph{ICCV}.

\bibitem[{Oord, Li, and Vinyals(2018)}]{oord2018representation}
Oord, A. v.~d.; Li, Y.; and Vinyals, O. 2018.
\newblock Representation learning with contrastive predictive coding.
\newblock \emph{arXiv preprint arXiv:1807.03748}.

\bibitem[{Paszke et~al.(2019)Paszke, Gross, Massa, Lerer, Bradbury, Chanan,
  Killeen, Lin, Gimelshein, Antiga et~al.}]{paszke2019pytorch}
Paszke, A.; Gross, S.; Massa, F.; Lerer, A.; Bradbury, J.; Chanan, G.; Killeen,
  T.; Lin, Z.; Gimelshein, N.; Antiga, L.; et~al. 2019.
\newblock Pytorch: An imperative style, high-performance deep learning library.
\newblock In \emph{NeurIPS}.

\bibitem[{Poole et~al.(2019)Poole, Ozair, Van Den~Oord, Alemi, and
  Tucker}]{poole2019variational}
Poole, B.; Ozair, S.; Van Den~Oord, A.; Alemi, A.; and Tucker, G. 2019.
\newblock On variational bounds of mutual information.
\newblock In \emph{ICML}.

\bibitem[{Qian and Cheung(2019)}]{qian2019enhancing}
Qian, D.; and Cheung, W.~K. 2019.
\newblock Enhancing variational autoencoders with mutual information neural
  estimation for text generation.
\newblock In \emph{EMNLP}.

\bibitem[{Ragonesi et~al.(2021)Ragonesi, Volpi, Cavazza, and
  Murino}]{ragonesi2021learning}
Ragonesi, R.; Volpi, R.; Cavazza, J.; and Murino, V. 2021.
\newblock Learning unbiased representations via mutual information
  backpropagation.
\newblock In \emph{CVPR Workshop}.

\bibitem[{Sagawa et~al.(2020{\natexlab{a}})Sagawa, Koh, Hashimoto, and
  Liang}]{GroupDRO}
Sagawa, S.; Koh, P.~W.; Hashimoto, T.~B.; and Liang, P. 2020{\natexlab{a}}.
\newblock Distributionally robust neural networks for group shifts: On the
  importance of regularization for worst-case generalization.
\newblock In \emph{ICLR}.

\bibitem[{Sagawa et~al.(2020{\natexlab{b}})Sagawa, Raghunathan, Koh, and
  Liang}]{sagawa2020investigation}
Sagawa, S.; Raghunathan, A.; Koh, P.~W.; and Liang, P. 2020{\natexlab{b}}.
\newblock An Investigation of Why Overparameterization Exacerbates Spurious
  Correlations.
\newblock In \emph{ICML}.

\bibitem[{Seo, Lee, and Han(2021)}]{seo2021unsupervised}
Seo, S.; Lee, J.-Y.; and Han, B. 2021.
\newblock Unsupervised Learning of Debiased Representations with
  Pseudo-Attributes.
\newblock \emph{arXiv preprint arXiv:2108.02943}.

\bibitem[{Sohoni et~al.(2020)Sohoni, Dunnmon, Angus, Gu, and
  R\'{e}}]{sohoni2020no}
Sohoni, N.; Dunnmon, J.; Angus, G.; Gu, A.; and R\'{e}, C. 2020.
\newblock No Subclass Left Behind: Fine-Grained Robustness in Coarse-Grained
  Classification Problems.
\newblock In \emph{NeurIPS}.

\bibitem[{Spirtes et~al.(2000)Spirtes, Glymour, Scheines, and
  Heckerman}]{spirtes2000causation}
Spirtes, P.; Glymour, C.~N.; Scheines, R.; and Heckerman, D. 2000.
\newblock \emph{Causation, prediction, and search}.
\newblock MIT press.

\bibitem[{Sun et~al.(2020)Sun, Hoffmann, Verma, and Tang}]{sun2019infograph}
Sun, F.-Y.; Hoffmann, J.; Verma, V.; and Tang, J. 2020.
\newblock Infograph: Unsupervised and semi-supervised graph-level
  representation learning via mutual information maximization.
\newblock In \emph{ICLR}.

\bibitem[{Suzuki et~al.(2008)Suzuki, Sugiyama, Sese, and
  Kanamori}]{suzuki2008approximating}
Suzuki, T.; Sugiyama, M.; Sese, J.; and Kanamori, T. 2008.
\newblock Approximating mutual information by maximum likelihood density ratio
  estimation.
\newblock In \emph{PMLR}.

\bibitem[{Thomee et~al.(2016)Thomee, Shamma, Friedland, Elizalde, Ni, Poland,
  Borth, and Li}]{thomee2016yfcc100m}
Thomee, B.; Shamma, D.~A.; Friedland, G.; Elizalde, B.; Ni, K.; Poland, D.;
  Borth, D.; and Li, L.-J. 2016.
\newblock YFCC100M: The new data in multimedia research.
\newblock \emph{Communications of the ACM}, 59(2): 64--73.

\bibitem[{Tishby, Pereira, and Bialek(2000)}]{tishby2000information}
Tishby, N.; Pereira, F.~C.; and Bialek, W. 2000.
\newblock The information bottleneck method.
\newblock \emph{arXiv preprint physics/0004057}.

\bibitem[{Van~der Maaten and Hinton(2008)}]{van2008visualizing}
Van~der Maaten, L.; and Hinton, G. 2008.
\newblock {Visualizing data using t-SNE.}
\newblock \emph{Journal of machine learning research}, 9(11).

\bibitem[{Wah et~al.(2011)Wah, Branson, Welinder, Perona, and
  Belongie}]{wah2011caltech}
Wah, C.; Branson, S.; Welinder, P.; Perona, P.; and Belongie, S. 2011.
\newblock The caltech-ucsd birds-200-2011 dataset.

\bibitem[{Wang et~al.(2019)Wang, Zhao, Yatskar, Chang, and
  Ordonez}]{wang2019iccv}
Wang, T.; Zhao, J.; Yatskar, M.; Chang, K.-W.; and Ordonez, V. 2019.
\newblock Balanced Datasets Are Not Enough: Estimating and Mitigating Gender
  Bias in Deep Image Representations.
\newblock In \emph{ICCV}.

\bibitem[{Wang et~al.(2020)Wang, Qinami, Karakozis, Genova, Nair, Hata, and
  Russakovsky}]{wang2020towards}
Wang, Z.; Qinami, K.; Karakozis, I.~C.; Genova, K.; Nair, P.; Hata, K.; and
  Russakovsky, O. 2020.
\newblock Towards fairness in visual recognition: Effective strategies for bias
  mitigation.
\newblock In \emph{CVPR}.

\bibitem[{Woodward(2005)}]{woodward2005making}
Woodward, J. 2005.
\newblock \emph{Making things happen: A theory of causal explanation}.
\newblock Oxford university press.

\bibitem[{Xie et~al.(2016)Xie, Wang, Wei, Wang, and Tian}]{xie2016disturblabel}
Xie, L.; Wang, J.; Wei, Z.; Wang, M.; and Tian, Q. 2016.
\newblock Disturblabel: Regularizing cnn on the loss layer.
\newblock In \emph{CVPR}.

\bibitem[{Zhang, Lemoine, and Mitchell(2018)}]{zhang2018mitigating}
Zhang, B.~H.; Lemoine, B.; and Mitchell, M. 2018.
\newblock Mitigating unwanted biases with adversarial learning.
\newblock In \emph{AAAI}.

\bibitem[{Zhang et~al.(2020)Zhang, Marklund, Gupta, Levine, and
  Finn}]{zhang2020adaptive}
Zhang, M.; Marklund, H.; Gupta, A.; Levine, S.; and Finn, C. 2020.
\newblock Adaptive Risk Minimization: A Meta-Learning Approach for Tackling
  Group Shift.
\newblock \emph{arXiv preprint arXiv:2007.02931}.

\bibitem[{Zhang and Sang(2020)}]{zhang2020towards}
Zhang, Y.; and Sang, J. 2020.
\newblock Towards Accuracy-Fairness Paradox: Adversarial Example-based Data
  Augmentation for Visual Debiasing.
\newblock \emph{arXiv preprint arXiv:2007.13632}.

\bibitem[{Zhao et~al.(2017)Zhao, Wang, Yatskar, Ordonez, and
  Chang}]{zhao2017men}
Zhao, J.; Wang, T.; Yatskar, M.; Ordonez, V.; and Chang, K.-W. 2017.
\newblock Men also like shopping: Reducing gender bias amplification using
  corpus-level constraints.
\newblock \emph{arXiv preprint arXiv:1707.09457}.

\bibitem[{Zhou et~al.(2017)Zhou, Lapedriza, Khosla, Oliva, and
  Torralba}]{zhou2017places}
Zhou, B.; Lapedriza, A.; Khosla, A.; Oliva, A.; and Torralba, A. 2017.
\newblock Places: A 10 million image database for scene recognition.
\newblock \emph{TPAMI}, 40(6): 1452--1464.

\end{thebibliography}
}

\end{document}